\documentclass[preprint]{jpsj2}

\usepackage{latexsym}
\usepackage{bm}
\usepackage{graphicx}

\def\gsize{0.50} 
\def\gs2{0.70} 

\title{Statistical Mechanics of On-line Learning 
when a Moving Teacher Goes around an Unlearnable True Teacher}

\author{
\textsc{Masahiro URAKAMI}$^{1}$,
\textsc{Seiji MIYOSHI}$^{2}$
\thanks{E-mail address: miyoshi@kobe-kosen.ac.jp}
and 
\textsc{Masato OKADA}$^{3}$,$^{4}$
}

\inst{
$^{1}$Department of Mathematics, 
Faculty of Science,
Kobe University, \\
1--1, Rokkodai, Nada-ku, Kobe 657--8501\\
$^{2}$Department of Electronic Engineering, 
Kobe City College of Technology, \\
8--3 Gakuen-higashimachi, Nishi-ku, Kobe, 651--2194 \\
$^{3}$Division of Transdisciplinary Sciences, 
Graduate School of Frontier Sciences, \\
The University of Tokyo, 
5--1--5 Kashiwanoha, Kashiwa-shi, Chiba, 277--8561 \\
$^{4}$RIKEN Brain Science Institute, 
2--1 Hirosawa, Wako-shi, Saitama, 351--0198
}

\abst{
In the framework of on-line learning,
a learning machine might move around a teacher
due to the differences in structures or output functions
between the teacher and the learning machine.
In this paper we analyze the generalization performance 
of a new student supervised by a moving 
machine.
A model composed of a fixed true teacher,
a moving teacher, and a student 
is treated theoretically
using statistical mechanics, 
where the true teacher is a nonmonotonic perceptron
and the others are simple perceptrons.
Calculating the generalization errors numerically,
we show that the generalization errors of a student
can temporarily become smaller than that of a moving teacher,
even if the student only uses examples 
from the moving teacher.
However, the generalization error of the student 
eventually becomes the same value with that of the moving teacher.
This behavior is qualitatively 
different from that of a linear model.
}

\kword{
on-line learning, generalization error,
moving teacher, true teacher, unlearnable case, 
nonmonotonic teacher
}

\begin{document}
\maketitle


\section{Introduction}
Learning is to infer the underlying rules that dominate 
data generation using observed data.
The observed data are input-output pairs from a teacher
and are called examples.
Learning can be roughly classified into batch learning and
on-line learning \cite{Saad}.
In batch learning, some given examples are used more than once,
a paradigm in which a student comes to give correct answers
after training if that student has an adequate degree of freedom.
However, it is necessary to have a long amount of time and 
a large memory in which many examples
may be stored.
On the contrary, examples used once are discarded
in on-line learning.
In this case, a student cannot give correct answers 
for all examples used in training.
However, there are some merits: for example,
a large memory for storing many examples is not necessary
and 
it is possible to follow a time-variant teacher. 

Recently, we \cite{Hara,PRE} 
have analyzed the generalization performance
of ensemble learning
in a framework of on-line learning
using a statistical mechanical method \cite{Saad,NishimoriE}.
In that process, the following points
were proven subsidiarily:
The generalization error does not approach zero
when the student is a simple perceptron and the teacher
is a committee machine \cite{IBIS2004} 
or a non-monotonic perceptron \cite{NC200503}.
Therefore, models like these can be called 
unlearnable cases \cite{Inoue,Inoue2,Inoue3}.
The behavior of a student in an unlearnable case
depends on the learning rule.
That is, the student vector asymptotically
converges in one direction using Hebbian learning.
On the contrary, the student vector 
does not converge in one direction but continues moving
using perceptron learning or AdaTron learning.
In the case of a non-monotonic teacher,
the student's behavior can be expressed by continuing to 
go around the teacher, 
keeping a constant direction cosine with the teacher.

Considering the applications of statistical learning theories,
investigating the 
system behaviors of unlearnable cases
is significant
since real-world problems seem to 
include many unlearnable cases.
In addition, 
a learning machine may continue going around a teacher
in the unlearnable cases as mentioned above.
Here, let us consider a new student that
is supervised by a moving learning machine.
That is, we consider a student that
uses the input-output pairs of 
a moving teacher as training examples,
and we investigate the generalization performance 
of a student for a true teacher.
Here, the true teacher is fixed.
Note that the examples used by the student
are only from the moving teacher, 
and the student cannot directly observe the outputs
of the true teacher.
In a real human society, a teacher
that can be observed by a student
does not always present the correct answer;
in many cases, the teacher is learning
and continues to vary.
Therefore, analyzing such a model
is interesting for considering the 
analogies between statistical learning theories 
and a real society.

A model 
in which a true teacher, a moving teacher, and a student
are all linear perceptrons \cite{Hara} with noises
was already solved analytically\cite{moveteacher}.
It was proved that
a student's generalization errors
can be smaller than that of the moving teacher
in the linear case
even though the student uses 
only the examples of the moving teacher.
However, linear perceptrons are somewhat special
as neural networks or learning machines.
Nonlinear perceptrons are more common than 
linear ones.
Therefore, in this paper 
we treat a model in which a true teacher,
a moving teacher, and a student are all nonlinear 
perceptrons.
We calculate the order parameters and 
the generalization errors
in the case of a true teacher as nonmonotonic
while the others are simple perceptrons
theoretically using a statistical mechanical method 
in the framework of on-line learning.
As a result, it is proved that
a student's generalization errors
can be smaller than that of the moving teacher.
That means the student can be cleverer than the moving teacher
even though the student uses only the examples from 
the moving teacher.
Although these behaviors are analogous to those of a linear model,
the generalization error of the student 
eventually becomes the same value as 
that of the moving teacher in the nonlinear model.

\section{Model}
Three nonlinear perceptrons are treated in this paper:
a true teacher, a moving teacher and a student.
Their connection weights are
$\mbox{\boldmath $A$},\mbox{\boldmath $B$}$, and
$\mbox{\boldmath $J$}$, respectively.
For simplicity, the connection weights of the true teacher,
that of the moving teacher and that of the student
are simply called the true teacher, the moving teacher, and
the student, respectively.
The true teacher 
$\mbox{\boldmath $A$}=(A_1,\ldots,A_N)$,
the moving teacher
$\mbox{\boldmath $B$}=(B_1,\ldots,B_N)$,
the student
$\mbox{\boldmath $J$}=(J_1,\ldots,J_N)$, and
input $\mbox{\boldmath $x$}=(x_1,\ldots,x_N)$
are $N$-dimensional vectors.
Each component $A_i$ of $\mbox{\boldmath $A$}$
is drawn from ${\cal N}(0,1)$ independently and fixed,
where ${\cal N}(0,1)$ denotes the Gaussian distribution with
a mean of zero and a variance of unity.
Each of the components $B_i^0, J_i^0$
of the initial values of 
$\mbox{\boldmath $B$},\mbox{\boldmath $J$}$
are drawn from ${\cal N}(0,1)$ independently.
Each component $x_i$ of $\mbox{\boldmath $x$}$
is drawn from ${\cal N}(0,1/N)$ independently.
Thus,
\begin{align}
\left\langle A_i\right\rangle &= 0, &
\left\langle (A_i)^2\right\rangle&=1, \\
\left\langle B_i^0\right\rangle &= 0, &
\left\langle (B_i^0)^2\right\rangle&=1,\\
\left\langle J_i^0\right\rangle &= 0, &
\left\langle (J_i^0)^2\right\rangle&=1,\\
\left\langle x_i\right\rangle &= 0, &
\left\langle (x_i)^2\right\rangle&=\frac{1}{N},
\end{align}
where $\langle \cdot \rangle$ denotes a mean.

In this paper, the thermodynamic limit $N\rightarrow \infty$
is also treated. Therefore,
\begin{equation}
\|\mbox{\boldmath $A$}\|=\sqrt{N},\ \ 
\|\mbox{\boldmath $B$}^0\|=\sqrt{N},\ \ 
\|\mbox{\boldmath $J$}^0\|=\sqrt{N},\ \ 
\|\mbox{\boldmath $x$}\|=1,
\label{eqn:xBJ}
\end{equation}
where $\|\cdot \|$ denotes a vector norm.
Generally, norms $\|\mbox{\boldmath $B$}\|$ and $\|\mbox{\boldmath $J$}\|$
of the moving teacher and the student
change as the time step proceeds.
Therefore, the ratios $l_B$ and $l_J$ of the norms to $\sqrt{N}$
are introduced and are called the length of the moving teacher
and the length of the student. That is,
$\|\mbox{\boldmath $B$}\|=l_B\sqrt{N}$，
$\|\mbox{\boldmath $J$}\|=l_J\sqrt{N}$.

The internal potentials $y$ of the true teacher, 
$vl_B$ of the moving teacher, 
and $ul_J$ of the student are
\begin{eqnarray}
y&=&\mbox{\boldmath $A$} \cdot\mbox{\boldmath $x$},\\
vl_B&=&\mbox{\boldmath $B$} \cdot\mbox{\boldmath $x$},\\
ul_J&=&\mbox{\boldmath $J$} \cdot\mbox{\boldmath $x$},
\end{eqnarray}
where $y$, $v$, and $u$
obey the Gaussian distributions with means of zero and 
variances of unity.

The output of the true teacher, 
which has a nonmonotonic output function,
is
\begin{equation}
d=\mbox{sgn}((y-a)y(y+a)),
\end{equation}
where $a$ is a fixed threshold of the nonmonotonic function.
The outputs of the moving teacher and the student, 
which are simple perceptrons, 
are $\mbox{sgn}(vl_B)$ and $\mbox{sgn}(ul_J)$, respectively.
Here, $\mbox{sgn}(\cdot)$ is a sign function defined as
\begin{eqnarray}
\mbox{sgn}(z)
&=&\left\{
\begin{array}{ll}
+1,            & z \geq 0 , \\
-1,            & z <    0 .
\end{array}
\right.
\label{eqn:sgn}
\end{eqnarray}

In the model treated in this paper,
the moving teacher $\mbox{\boldmath $B$}$
is updated using an input $\mbox{\boldmath $x$}$
and an output of the true teacher $\mbox{\boldmath $A$}$
for the input $\mbox{\boldmath $x$}$.
The student $\mbox{\boldmath $J$}$
is updated using an input $\mbox{\boldmath $x$}$
and an output of the moving teacher $\mbox{\boldmath $B$}$
for the input $\mbox{\boldmath $x$}$.
The moving teacher is considered to
use perceptron learning.
That is,
\begin{eqnarray}
\mbox{\boldmath $B$}^{m+1}
&=&\mbox{\boldmath $B$}^{m}+
\eta_B\Theta (-v^m d^m)d^m \mbox{\boldmath $x$}^m \\
&=&\mbox{\boldmath $B$}^{m}+
\eta_B\Theta (-v^m(y^m-a)y^m(y^m+a))\mbox{sgn}((y^m-a)y^m(y^m+a))
\mbox{\boldmath $x$}^m,
\label{eqn:updateB}
\end{eqnarray}
where $\eta_B$ denotes the learning rate
of the moving teacher and is a constant number.
Furthermore, $m$ denotes the time step, and
$\Theta(\cdot)$ denotes the step function defined as
\begin{eqnarray}
\Theta (z)
&=&\left\{
\begin{array}{ll}
+1,           & z \geq 0 , \\
0,            & z <    0 .
\end{array}
\right.\label{eqn:Theta}
\end{eqnarray}

The student is also considered to
use perceptron learning.
That is,
\begin{eqnarray}
\mbox{\boldmath $J$}^{m+1}
&=&\mbox{\boldmath $J$}^{m}+
\eta_J\Theta (-u^mv^m)\mbox{sgn}(v^m)\mbox{\boldmath $x$}^m,
\label{eqn:updateJ}
\end{eqnarray}
where $\eta_J$ denotes the student's learning rate
and is a constant number.
Generalizing the learning rules, 
Eqs. (\ref{eqn:updateB}) and (\ref{eqn:updateJ})
can be expressed as
\begin{eqnarray}
\mbox{\boldmath $B$}^{m+1}
&=&\mbox{\boldmath $B$}^{m}+g^{m}\mbox{\boldmath $x$}^{m},\\
\mbox{\boldmath $J$}^{m+1}
&=&\mbox{\boldmath $J$}^{m}+f^{m}\mbox{\boldmath $x$}^{m},
\end{eqnarray}
respectively.
Here, $g$ and $f$ are update functions of 
the moving teacher and the student, respectively.

\section{Theory}
\subsection{Generalization Error}
\begin{figure}[htbp]
\begin{center}
\includegraphics[width=\gsize\linewidth,keepaspectratio]{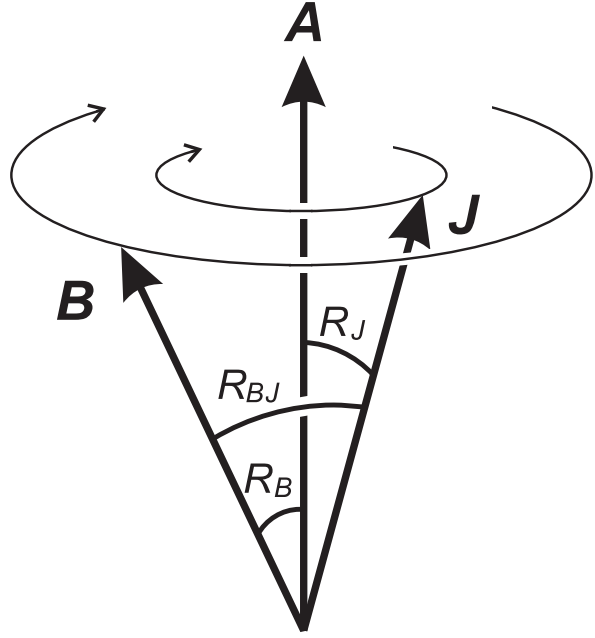}
\caption{True teacher $\mbox{\boldmath $A$}$,
moving teacher $\mbox{\boldmath $B$}$, and 
student $\mbox{\boldmath $J$}$.
$R_B, R_J$, and $R_{BJ}$ are direction cosines.}
\label{fig:ABJ}
\end{center}
\end{figure}

A goal of a statistical learning theory
is to theoretically obtain generalization errors.
We use 
\begin{eqnarray}
\epsilon_B^m 
&=&
\Theta \left(-
d^m
\mbox{sgn}(v^ml_B^m)\right) \\
&=&
\Theta \left(-
(y^m-a)y^m(y^m+a)
v^m \right)
\label{eqn:eB}
\end{eqnarray}
and 
\begin{eqnarray}
\epsilon_J^m 
&=&
\Theta \left(-
d^m
\mbox{sgn}(u^ml_J^m)
\right) \\
&=&
\Theta \left(-
(y^m-a)y^m(y^m+a)
u^m \right)
\label{eqn:eJ}
\end{eqnarray}
as errors of the moving teacher and student,
respectively.
The superscripts $m$, which represent the time steps, 
are omitted for simplicity.
We define a generalization error 
as a mean of error over the distribution
$p(\mbox{\boldmath $x$})$
of inputs $\mbox{\boldmath $x$}$.
The error $\epsilon_B$ of the moving teacher 
and the error $\epsilon_J$ of the student 
can be expressed as 
$\epsilon_B(y,v)$ and $\epsilon_J(y,u)$
using $y, v$, and $u$,
Therefore, 
the generalization error $\epsilon_{gB}$
of the moving teacher
and 
the generalization error $\epsilon_{gJ}$
of the student
can be calculated using 
the distributions $p(y,v)$ and $p(y,u)$ as follows:
\begin{eqnarray}
\epsilon_{gB} 
&=& \langle \epsilon_B \rangle_{\bm{x}} \\
&=& \int d\mbox{\boldmath $x$}p(\mbox{\boldmath $x$})\epsilon_B \\ 
&=& \int dydvp(y,v)\epsilon_B(y,v), 
\label{eqn:egB}\\
\epsilon_{gJ} 
&=& \langle \epsilon_J \rangle_{\bm{x}} \\
&=& \int d\mbox{\boldmath $x$}p(\mbox{\boldmath $x$})\epsilon_J \\
&=& \int dydup(y,u)\epsilon_J(y,u). 
\label{eqn:egJ}
\end{eqnarray}
Since $y,v$ and $u$
are calculated using 
$\mbox{\boldmath $A$}, \mbox{\boldmath $B$}, \mbox{\boldmath $J$}$,
and the independent input $\mbox{\boldmath $x$}$,
$p(y,v,u)$ is the multiple Gaussian distribution
with means of zero and the 
covariance matrix $\mbox{\boldmath $\Sigma$}$
\begin{eqnarray}
\mbox{\boldmath $\Sigma$}
  &=&
   \left(
   \arraycolsep=3pt
   \begin{array}{ccc}
     1       & R_B     & R_J       \\
     R_B     &   1     & R_{BJ}    \\
     R_J     & R_{BJ}  &    1  
   \end{array}
   \right).\label{eqn:Sigma}
\end{eqnarray}
Here, 
$R_B$ is the direction cosine between 
$\mbox{\boldmath $A$}$ and $\mbox{\boldmath $B$}$.
$R_J$ is the direction cosine between 
$\mbox{\boldmath $A$}$ and $\mbox{\boldmath $J$}$.
$R_{BJ}$ is the direction cosine between 
$\mbox{\boldmath $B$}$ and $\mbox{\boldmath $J$}$.
Thus,
\begin{eqnarray}
R_B&=&\frac{\mbox{\boldmath $A$}\cdot\mbox{\boldmath $B$}} 
{\|\mbox{\boldmath $A$}\|\| \mbox{\boldmath $B$}\| },\\
R_J&=&\frac{\mbox{\boldmath $A$}\cdot\mbox{\boldmath $J$} } 
{\|\mbox{\boldmath $A$}\|\| \mbox{\boldmath $J$}\| },\\
R_{BJ}&=&\frac{\mbox{\boldmath $B$}\cdot\mbox{\boldmath $J$} } 
{\|\mbox{\boldmath $B$}\|\| \mbox{\boldmath $J$}\| }.
\end{eqnarray}
Equations (\ref{eqn:egB}) and (\ref{eqn:egJ}) 
can be calculated by excuting the Gaussian integrations
using these direction cosines 
as follows\cite{Inoue,Inoue2,Inoue3}:
\begin{eqnarray}
\epsilon_{gB} 
&=& \int_{0}^{a}DyH\left(\frac{-yR_B}{\sqrt{1-R_B^2}}\right) 
+ \int_{\infty}^{-a}DyH\left(\frac{-yR_B}{\sqrt{1-R_B^2}}\right),
\label{eqn:egB2}\\
\epsilon_{gJ} 
&=& \int_{0}^{a}DyH\left(\frac{-yR_J}{\sqrt{1-R_J^2}}\right) 
+ \int_{\infty}^{-a}DyH\left(\frac{-yR_J}{\sqrt{1-R_J^2}}\right),
\label{eqn:egJ2}
\end{eqnarray}
where
\begin{equation}
H(u) \equiv \int_u^\infty Dy,\ \ \ 
Dy \equiv \frac{dy}{\sqrt{2\pi}}\exp\left(-\frac{y^2}{2}\right).
\label{eqn:Dz}
\end{equation}
The relationship among the true teacher $\mbox{\boldmath $A$}$,
the moving teacher $\mbox{\boldmath $B$}$,
and the student $\mbox{\boldmath $J$}$
is shown in Fig. \ref{fig:ABJ}.

\subsection{Differential equations of order parameters}
Since we treat the thermodynamic limit $N\rightarrow \infty$
in this paper,
$O(N)$ updates of 
Eqs. (\ref{eqn:updateB}) and (\ref{eqn:updateJ})
are necessary for the order parameters to change $O(1)$.
Therefore, we denote time steps $m$ normalized by 
the dimension $N$ as a continuous time $t=m/N$.
We use $t$ as a subscript for the learning process.

The generalization errors
$\epsilon_{gB}$ and $\epsilon_{gJ}$
can be calculated if all the order parameters 
$R_B$, $R_J$ and $R_{BJ}$ are known.
Therefore, 
simultaneous differential equations
in deterministic forms \cite{NishimoriE} 
have been obtained
that describe
the dynamical behaviors of order parameters
based on self-averaging
in the thermodynamic limits as follows\cite{moveteacher}:
\begin{align}
\frac{dl_B}{dt}
&=
\langle gv \rangle + \frac{\langle g^2 \rangle}{2l_B},\label{difeqn5}
\\
\frac{dl_J}{dt}
&=
\langle fu \rangle + \frac{\langle f^2 \rangle}{2l_J},\label{difeqn4}\\
\frac{dR_{BJ}}{dt}
&=
-R_{BJ}\left(\frac{1}{l_J}\frac{dl_J}{dt}+\frac{1}{l_B}\frac{dl_B}{dt}\right)+\frac{1}{l_B}\langle gu\rangle 
+\frac{1}{l_J}\langle fv\rangle 
+\frac{1}{l_Bl_J}\langle gf\rangle, \label{difeqn1} \\
\frac{dR_J}{dt}
&=
\frac{1}{l_J}
\left(-R_J\frac{dl_J}{dt}+\langle fy \rangle\right),\label{difeqn2}\\
\frac{dR_B}{dt}
&=
\frac{\langle gy\rangle -\langle gv\rangle R_B}{l_B}
-\frac{R_B}{2l_B^2}\langle gy\rangle. \label{difeqn3}
\end{align}

As mentioned above, $y$, $v$, and $u$
obey the triple Gaussian distribution
with means of zero and the covariance matrix
of Eq. (\ref{eqn:Sigma}).
Using this, we can calculate
the nine sample averages that appear in 
Eqs. (\ref{difeqn5})--(\ref{difeqn3}) as follows:


\begin{align} 
\langle gv \rangle 
&=\frac{\eta_B}{\sqrt{2\pi}}
\left(R_B\left(2\exp
\left(-\frac{a^2}{2}\right)-1\right)-1\right)
\label{gv}
,\\
\langle g^2 \rangle 
&=2{\eta}_B^2\left(\int_{0}^{a}D_yH\left(\frac{-yR_B}{\sqrt{1-R_B^2}}\right) 
+ \int_{\infty}^{-a}D_yH\left(\frac{-yR_B}{\sqrt{1-R_B^2}}\right)\right)
,\\
\langle fu \rangle 
&={\eta}_J\frac{R_{BJ}-1}{\sqrt{2\pi}}
,\\
\langle f^2 \rangle 
&=\frac{{\eta}_J^2}{\pi}{\tan}^{-1}\frac{\sqrt{1-R_{BJ}^2}}{R_{BJ}}
,\\
\langle gu \rangle 
&=\frac{\eta_B}{\sqrt{2\pi}}
\left(R_J\left(2\exp
\left(-\frac{a^2}{2}\right)-1\right)-R_{BJ}\right)
\label{gu}
,\\
\langle fv \rangle 
&={\eta}_J\frac{1-R_{BJ}}{\sqrt{2\pi}}
,\\
\langle gf \rangle 
&=-2{\eta}_B{\eta}_J\left( \int_{0}^{a} D_y+\int_{-\infty}^{a} D_y\right) 
\int_{-\frac{yR_B}{\sqrt{1-R_B^2}}}^{\infty}Dv
H\left( -\frac{yR_J\sqrt{1-R_B^2}+v(R_BR_J-R_{BJ})}{\sqrt{1+2R_BR_JR_{BJ}-
R_B^2-R_J^2-R_{BJ}^2}}\right) \label{gf} 
,\\
\langle fy \rangle 
&={\eta}_J\frac{R_B-R_J}{\sqrt{2\pi}} \label{fy} 
,\\
\langle gy \rangle 
&=\frac{{\eta}_B}{\sqrt{2\pi}}\left(2\exp\left(-\frac{a^2}{2}\right)-
1-R_B\right).
\label{gy}
\end{align}


\section{Results and discussion}
\begin{figure}[htbp]
\begin{center}
\includegraphics[width=\gs2\linewidth,keepaspectratio]{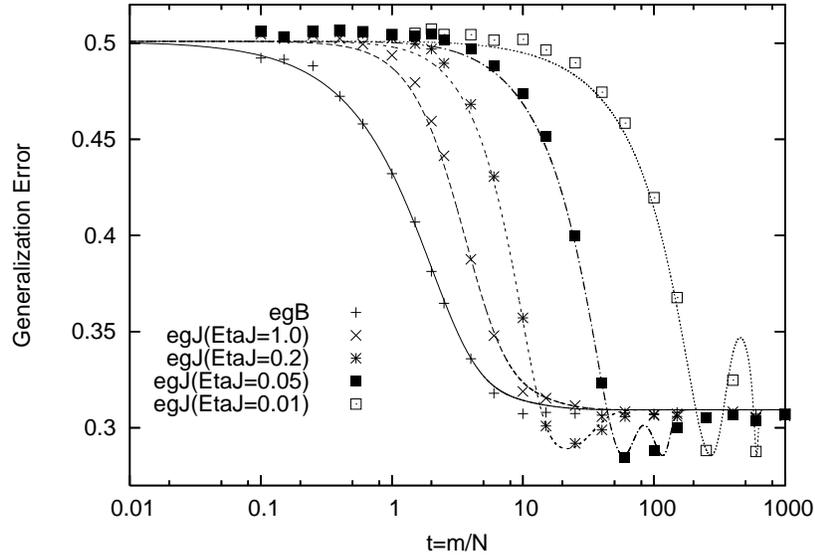}
\end{center}
\caption{Dynamical behaviors of $\epsilon_{gB}$ and $\epsilon_{gJ}$.
Conditions are $a=0.5$ and $\eta_B=0.1$.
Curves represent theoretical results
and symbols 
represent simulation results, where $N=10^4$.}
\label{eg}
\end{figure}

\begin{figure}[htbp]
\begin{center}
\includegraphics[width=\gs2\linewidth,keepaspectratio]{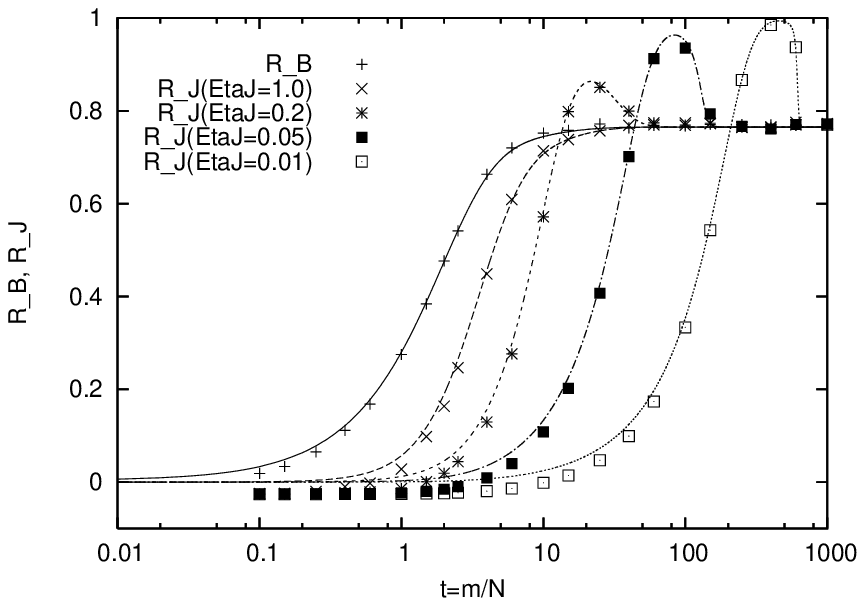}
\end{center}
\caption{Dynamical behaviors of $R_B$ and $R_J$.
Conditions are $a=0.5$ and $\eta_B=0.1$.
Curves represent theoretical results
and symbols 
represent simulation results, where $N=10^4$.}
\label{R}
\end{figure}

\begin{figure}[htbp]
\begin{center}
\includegraphics[width=\gs2\linewidth,keepaspectratio]{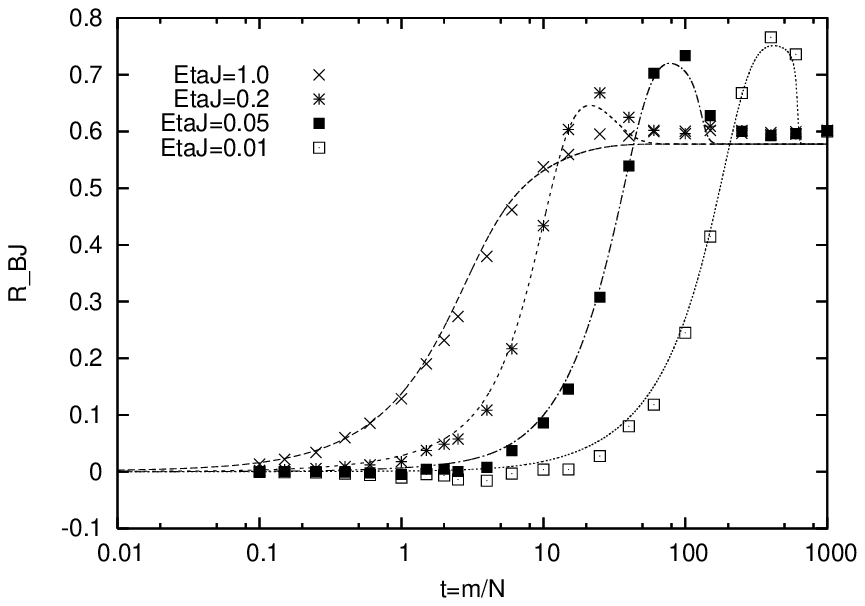}
\end{center}
\caption{Dynamical behaviors of $R_{BJ}$.
Conditions are $a=0.5$ and $\eta_B=0.1$.
Curves represent theoretical results
and symbols 
represent simulation results, where $N=10^4$.}
\label{RBJ}
\end{figure}

\begin{figure}[htbp]
\begin{center}
\includegraphics[width=\gs2\linewidth,keepaspectratio]{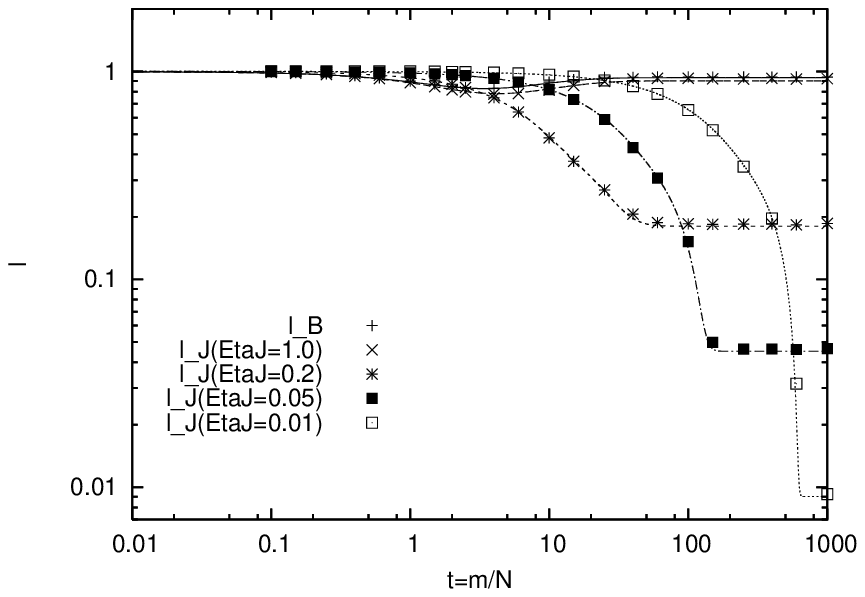}
\end{center}
\caption{Dynamical behaviors of $l_B$ and $l_J$.
Conditions are $a=0.5$ and $\eta_B=0.1$.
Curves represent theoretical results
and symbols 
represent simulation results, where $N=10^4$.}
\label{l}
\end{figure}

Figures \ref{eg}--\ref{l} illustrate the dynamical behaviors of 
the generalization errors and the order parameters.
The threshold $a$ of the true teacher is $0.5$
and the learning rate $\eta_B$ of the moving teacher
is $0.1$.
In these figures, 
the curves represent the theoretical results
and the symbols 
represent the simulation results, where $N=10^4$.
In theoretical calculations,
the simultaneous differential equations
have been solved numerically
using the sample averages in 
Eqs. (\ref{difeqn5})--(\ref{difeqn3})
also obtained numerically.
The generalization errors $\epsilon_{gB}$ and $\epsilon_{gJ}$
are calculated by executing integrations in 
Eqs. (\ref{eqn:egB2}) and (\ref{eqn:egJ2}) numerically
using the obtained $R_B, R_J$, and $R_{BJ}$.
In the computer simulations, 
the generalization errors 
have been measured through tests using $10^5$ random inputs
at each time step.
In these figures,
the theoretical results and the computer
simulations closely agree with each other.

Figure \ref{eg} shows that the student's
generalization error $\epsilon_{gJ}$
is always larger than 
$\epsilon_{gB}$ of the moving teacher
when the student's learning rate $\eta_J$
is relatively large, for example $\eta_J=1.0$.
In that case, $\epsilon_{gJ}$ approaches $\epsilon_{gB}$
asymptotically.
On the other hand, $\epsilon_{gJ}$ temporarily becomes smaller than
$\epsilon_{gB}$ 
when the learning rate $\eta_J$
is relatively small, for example $\eta_J=0.2, 0.05$ or $0.01$.
This is an interesting phenomenon since 
the student can temporarily become cleverer than the moving teacher
even though the student uses only the examples from 
the moving teacher.
This is the same as the linear case\cite{moveteacher}
whereby $\epsilon_{gJ}$ can become smaller than $\epsilon_{gB}$.
In the linear case\cite{moveteacher},
a small $\epsilon_{gJ}$ is maintained after
$\epsilon_{gJ}$ becomes smaller than $\epsilon_{gB}$.
However, $\epsilon_{gJ}$ returns to the same value as 
$\epsilon_{gB}$ in the nonlinear case 
treated in this paper.
This behavior is interesting since 
it is qualitatively different from the linear case.
In addition, the overshot of $\epsilon_{gJ}$ occurs 
only once when $\eta_J=0.2$.
On the other hand, $\epsilon_{gJ}$ swings three times 
when $\eta_J=0.05, 0.01$.

Figure \ref{R} shows that
$R_J$ temporarily becomes larger than $R_B$ when $\eta_J$ is small.
This means that 
$\mbox{\boldmath $J$}$ comes closer to $\mbox{\boldmath $A$}$
than $\mbox{\boldmath $B$}$.
Although the overshot of $R_J$ occurs only once,
$\epsilon_{gJ}$ swings three times when $\eta_J=0.05, 0.01$.
The reason for this difference can be understood as follows.
In the case of a nonmonotonic teacher, the relationship between
the generalization error $\epsilon_g$ and 
the direction cosine $R$ is Eq. (\ref{eqn:egB2}) or 
(\ref{eqn:egJ2})\cite{Inoue,Inoue2,Inoue3}.
In the case of $a<\sqrt{2\ln 2}=1.18$,
$\epsilon_g$ is not a monotonic function of $R$
and takes a minimum value when 
$R=\sqrt{(2\ln 2-a^2)/(2\ln 2)}$.
Since $a=0.5$ is treated in this section,
$\epsilon_g$ takes a minimum value when $R=0.905$.
The theoretical curves of $\eta_J=0.05, 0.01$ in Fig. \ref{R}
indicate that $R$ agrees with $0.905$ twice.
This phenomenon corresponds to the two local minima
in Fig. \ref{eg}.
On the other hand, $R$ does not reach $0.905$ when 
$\eta_J=0.2$. Therefore, the number of the minimum of 
$\epsilon_g$ is also only one.

Figure \ref{R} shows that the maximum value of $R_J$ 
is unity when $\eta_J=0.01$.
This is also a very interesting phenomenon
since the direction cosine between a teacher and 
a student does not reach unity when the student
learns the nonmonotonic teacher using perceptron
learning\cite{Inoue3}.

In addition, $R_B$ and $R_J$ agree with each other 
after enough time steps. 
However, the moving teacher and the student 
do not coincide with each other.
That is, $R_{BJ}$
is smaller than unity as shown in Fig. \ref{RBJ}.

\section{Conclusion}
In the framework of on-line learning,
a learning machine might move around a teacher
due to the differences in structures or output functions
between the teacher and the learning machine.
In this paper we analyzed the generalization performance 
of a new student supervised by a moving 
machine.
A model composed of a fixed true teacher,
a moving teacher, and a student 
was treated theoretically
using statistical mechanics, 
where the true teacher is a nonmonotonic perceptron
and the others are simple perceptrons.
Calculating the generalization errors numerically,
we have shown that a student's the generalization error
can temporarily become smaller than that of a moving teacher,
even if the student only uses examples 
from the moving teacher.
However, the student's generalization error 
eventually becomes 
the same value as that of the moving teacher.
This behavior is qualitatively different from that of a linear model.

\section*{Acknowledgments}
This research was partially supported by the Ministry of Education, 
Culture, Sports, Science, and Technology of Japan, 
with Grants-in-Aid for Scientific Research
15500151, 16500093, 18020007, 18079003 and 18500183.


\end{document}